# Tuning Learning Rates with the Cumulative-Learning Constant

By Nathan Faraj

**Abstract**

This paper introduces a novel method for optimizing learning rates in machine learning. A previously unrecognized proportionality between learning rates and dataset sizes is discovered, providing valuable insights into how dataset scale influences training dynamics. Additionally, a cumulative learning constant is identified, offering a framework for designing and optimizing advanced learning rate schedules. These findings have the potential to enhance training efficiency and performance across a wide range of machine learning applications.

## 1. Introduction

The learning rate is a deep learning hyperparameter that is very important for training models efficiently. There has been a lot of research done on optimizing learning rates including sophisticated methods for adaptive learning rates as well as alternating learning rates (Wu et al., 2019). However, the learning rate selection is arbitrary - An optimal learning rate is found without understanding due to the belief that the learning rate is too sensitive to changes in other hyperparameters within the model. However, this paper will present that the learning rate follows some strict proportionality laws. Previous work has been done to compare proportionality of learning rates in reference to batch size (Diego et al., 2020). Therefore, this paper will not be exploring how batch size affects the optimal learning rate.

### 1.1 Main Contributions

In this paper, we demonstrate that the optimal learning rate is inversely proportional to the size of the dataset. Notably, we show that this proportionality holds regardless of whether the dataset is expanded with new data or repeated data; for instance, doubling the number of epochs and doubling the dataset size are equivalent in this context. Additionally, we introduce the concept of a "cumulative learning constant," a value specific to a given model architecture. This constant can be calculated at smaller scales and used to determine the optimal learning rate for larger datasets. Furthermore, we explore how the cumulative learning constant establishes a relationship between various scheduled learning rates and a constant learning rate, enabling more efficient optimization of advanced learning rate schedules.

## 2. Related Work

Work is continuously being done to de-parameterize deep neural networks, as it would greatly improve computational efficiency and reduce the costs associated with hyperparameter tuning. One major focus has been on eliminating manual learning rate selection, as learning rate tuning remains one of the most sensitive and computationally expensive aspects of training deep learning models.

As stated before, related work has been done in establishing proportional relationships between learning rate and batch size for both SGD and Adam optimizers (Diego et al., 2020). Such research has been crucial in defining scaling laws that allow batch size adjustments to compensate for changes in learning rate without degrading convergence (Goyal et al., 2017). Additionally, theoretical work has examined how batch size influences gradient noise and optimization stability (Shallue et al., 2019), reinforcing the importance of proportional adjustments in learning rates.

Learning rate schedules have been shown to be critical for optimizing sophisticated models, such as speech recognition models (Senior et al., 2013). Studies have demonstrated that cyclical learning rates (Smith, 2017) and cosine annealing schedules (Loshchilov & Hutter, 2017) significantly improve training efficiency by adapting the step size dynamically throughout the training process. The use of adaptive learning rate methods, such as Hypergradient Descent (Baydin et al., 2017), allows for real-time adjustment of learning rates based on gradient feedback.

Beyond learning rate schedules, research has explored theoretical optimizations of convergence rates, particularly for Lipschitz functions (Defazio & Mishchenko, 2023). This research is crucial in determining the optimal way to decay learning rates to balance stability and convergence. Additionally, meta-learning techniques such as Learning to Optimize (L2O) (Andrychowicz et al., 2016) and reinforcement learning-based optimization (Li et al., 2017) have aimed to automate the tuning of hyperparameters, including learning rates. Furthermore, hyperparameter-free training approaches like AutoLR (Mackay et al., 2019) and Hyperband (Li et al., 2017) attempt to remove manual intervention in learning rate tuning.

A promising development in this direction is D-adaptation, which removes the need for manual learning rate tuning by dynamically adjusting the learning rate based on the distance traveled in parameter space (Defazio, A., & Mishchenko, K., 2023). However, D-adaptation is not computationally efficient for sufficiently large models, as the overhead of dynamic step-size computation can outweigh its benefits when training transformers or diffusion models. Alternative gradient-based adaptation approaches, such as reversible learning rate optimization (Maclaurin et al., 2015) and Bayesian learning rate adaptation (Wu et al., 2018), have also been proposed but remain computationally expensive.

While this work does not aim to fully de-parameterize scheduled learning rates, it provides an alternative theoretical framework by demonstrating fundamental relationships between constant and scheduled learning rates. These insights contribute to further optimizing learning rate selection and lay the groundwork for future research toward fully automated, parameter-free learning rate adaptation.

## 3. Experimentation and Contextualization

The data being used is the MNIST dataset. The training set size was initially set to 60000. The pixels have been normalized to values between -1 and 1. The input layer has 784 neurons and outputs to a hidden layer of 256 neurons. The hidden layer outputs to the output layer of 10 neurons, corresponding to the possible digits 0-9. There is a ReLu function after both the input layer and the hidden layer. The loss function is cross-entropy loss, and the optimizer used is the Adam optimizer. While the Adam optimizer is not ideal for testing learning rates initially, it will be observed that the methodology proposed works for stochastic gradient descents as well. The control group will be used on 10 epochs. Loss measurements are taken in an evaluation function, where the model is no longer being trained. All the data presented was obtained by taking 3 data points at each timestep and calculating their average.

### 3.1 Initial Observations

Initially, rough data was taken to approximate the optimal constant learning rate for the model discussed in 2.1. This graph can be seen in figure (1) below.

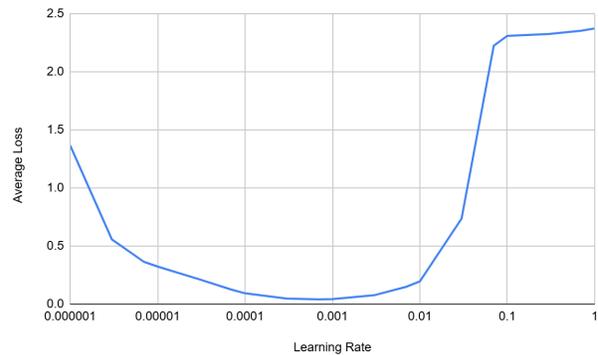

Figure 1. Average Loss vs. Learning Rate on 10 epochs.

Here, the loss seems to minimize at a learning rate right under 0.001. Initial experiments were performed to determine the effects that certain hyperparameters had on learning the optimal learning rates. The epochs were changed to 5, and new results were produced, seen in figure (2) below.

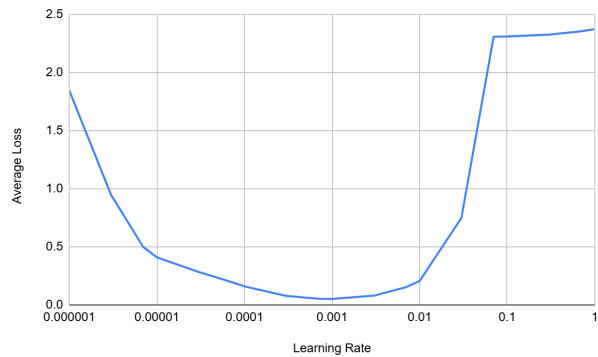

Figure 2. Average Loss vs. Learning Rate on 5 epochs using an Adam optimizer.

Here, the loss seemed to minimize at a learning rate of about 0.001, hinting that there may be a correlation between the number of epochs and the optimal learning rate. Next, the number of epochs was changed to 1. The results are shown in figure (3).

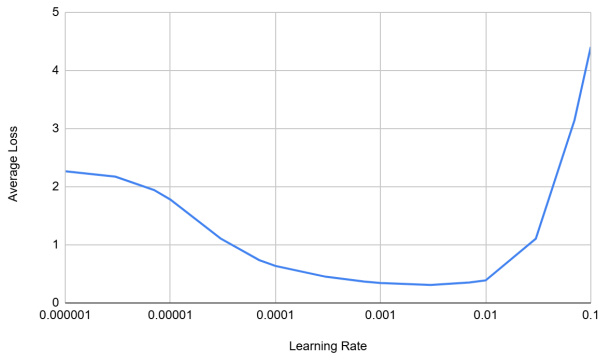

Figure 3. Average Loss vs. Learning Rate on 1 epoch using an Adam optimizer.

Here, the learning rate that minimized loss seemed to occur at about 0.003. In order to better understand what is happening, the number of epochs was set back to 10 and the data was halved instead. The results are in figure (4) below.

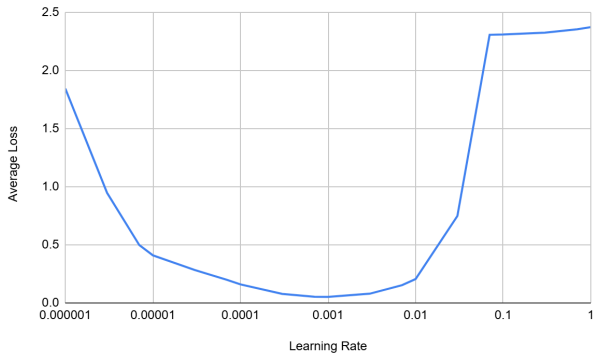

Figure 4. Average Loss vs. Learning Rate on a half-sized dataset using the Adam optimizer.

It appears from figure (4) that the optimal learning rate is identical to that shown in figure (2) where the number of epochs was halved. This hints at the idea that the total amount of data that a model is subject to is more important for optimizing the learning rate than the number of epochs or the amount of training data independently - they may be congruent to each other.

## 4. Thorough Analysis of Optimal Learning Rates

From here on forward, the training dataset was cut in half to 30,000 figures per epoch to save computational resources. More precise measurements were taken to find the optimal learning rate while varying the number of epochs between 1-80. Larger datasets are ideal, but the computational power required was not feasible. Regardless, there is a clear trend, which can be seen in figure (5).

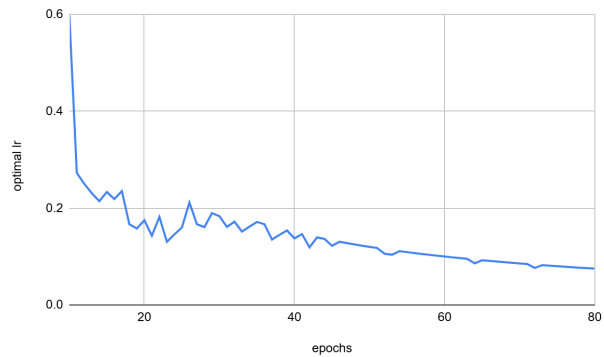

Figure 5. Optimal Learning Rates vs. Number of Epochs using an Adam Optimizer.

This experiment was also done with a SGD optimizer. The results can be seen in figure (6) below.

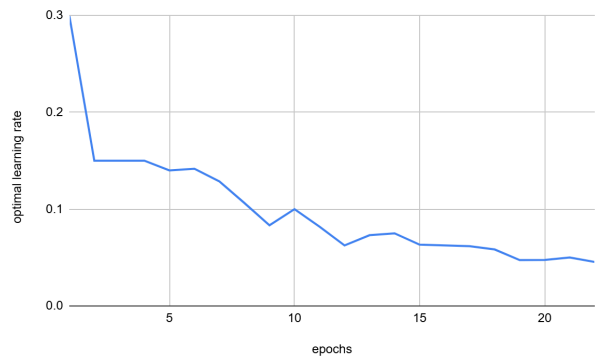

Figure 6. Optimal Learning Rates vs. Number of Epochs using the SGD Optimizer.

While the initial data at low epochs is choppy, it smooths out. We will compare data to develop a proportionality. Below is a table looking more closely at data obtained in figure (5).

Table 1. The table compares the number of epochs to the optimal learning rate under an Adam optimizer. It is set up such that it is easy to see the learning rate at epochs $n$ and epochs $2n$.

| Epochs | LR | Epochs | LR |
|---|---|---|---|
| 5 | 0.0011 | 10 | 0.00045 |
| 6 | 0.001167 | 12 | 0.000375 |
| 7 | 0.000786 | 14 | 0.000286 |
| 8 | 0.0008125 | 16 | 0.00025 |
| 9 | 0.000611 | 18 | 0.000333 |

It demonstrates that there is generally (but not quite) an inverse proportionality relating learning rates to the number of epochs. This insinuates that the Adam optimizer may play a role. On the contrary, the table below shows some data points using the SGD optimizer.

Table 2. The table compares the number of epochs to the optimal learning rate under an SGD optimizer. It is set up such that it is easy to see the learning rate at epochs $n$ and epochs $2n$.

| Epochs | LR | Epochs | LR |
|---|---|---|---|
| 5 | 0.14 | 10 | 0.1 |
| 6 | 0.1416666667 | 12 | 0.0625 |
| 7 | 0.1285714286 | 14 | 0.075 |
| 8 | 0.10625 | 16 | 0.0625 |
| 9 | 0.08333333333 | 18 | 0.05833333333 |

Here, it is clear that every time the number of epochs is doubled, the optimal learning rate halves.

**4.1 Inverse Proportionality Theorem**

It can be seen there is a clear trend between the number of epochs and the optimal learning rate. More precisely,

$$\eta \propto \frac{1}{k}$$

Where $\eta$ represents the change in optimal learning rate and $k$ represents the change in number of epochs. This is only valid for a constant learning rate.

**Proof:** The standard SGD update rule is:

$$x_t + 1 = x_t - \eta \nabla f(x_t)$$

For an L-smooth function, the function value after one step is bounded by:

$$f(x_{t+1}) \leq f(x_t) - \eta \|\nabla f(x_t)\|^2 + \frac{L}{2}\eta^2 \|\nabla f(x_t)\|^2$$

Taking expectations and summing over k steps:

$$E[f(x_k)] - f(x_0) \leq -\eta \sum_{t=0}^{k-1} E[\|\nabla f(x_t)\|^2] + \frac{L}{2}\eta^2 \sum_{t=0}^{k-1} E[\|\nabla f(x_t)\|^2]$$

Rearranging:

$$\eta \sum_{t=0}^{k-1} E[\|\nabla f(x_t)\|2] \leq f(x_0) - E[f(x_k)] + \frac{L}{2}\eta^2 \sum_{t=0}^{k-1} E[\|\nabla f(x_t)\|^2]$$

For sufficiently small $\eta$, the quadratic term is negligible, yielding:

$$\eta k E[\|\nabla f(x_t)\|^2] \leq C_1$$

where $C1 = f(x_0) - E[f(x_k)]$ is a problem-dependent constant. Solving for $\eta$:

$$\eta \leq \frac{C_1}{kE[\|\nabla f(x_t)\|^2]}$$

From standard SGD convergence results, for an optimal choice of $\eta$, the expected squared gradient norm satisfies:

$$E[\|\nabla f(x_t)\|^2] \approx O(1)$$

Thus, substituting this into the previous bound:

$$\eta_k = \frac{C}{k}$$

for some constant $C$ dependent on the function properties.

**4.2 Effect of Dataset Size on Learning Rates**

In order to better understand the role the size of a dataset plays in the learning rate, an experiment was performed to determine if dataset sizes behave exactly like epochs. For every increase in epoch, the dataset size was decreased proportionally. For instance, in going from epoch 1 to 2, the dataset size was cut in half. The experiment was performed this way in order to be computationally efficient. The results are seen in figure (7) below.

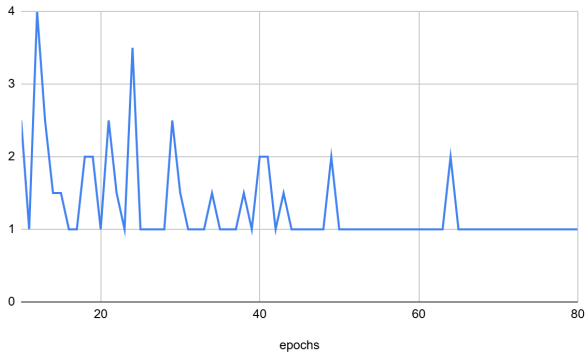

Figure 7. Optimal Learning Rate vs. Epochs while reducing dataset size proportionally to the increase in number of epochs. This experiment was performed using an Adam optimizer.

While choppy, the optimal learning rate is much more consistent than shown in figure (5). The conclusion to draw is that dataset sizes play the same role as epochs. The paper will refer to a term 'total data'. Precisely, this term is just the amount of data in the training dataset multiplied by the amount of epochs.

**Theorem 4.1.** The optimal learning rate is affected by the total amount of data a model will see during training.
**Theorem 4.2.** For a constant learning rate, the change in the optimal learning rate is inversely proportional to the change in the size of total data the model is trained on.

Therefore, the more precise proportionality is

$$\eta \propto \frac{1}{D}$$

Where η is the learning rate and $D$ is the total data. Again, this is a proportionality generalized for a constant learning rate only. The proof for this follows the same logic as proof 4.1 but it is generalized for all data instead of epochs specifically.

**5. Cumulative Learning Equation**

We will define an equation.

$$\kappa = \int_0^D \eta(x)dx$$

where κ represents the total learning effort in parameter space, $D$ denotes the total amount of data used in training, and $\eta(x)$ represents the learning rate at a given training step x. This integral reflects the total accumulated movement in weight space as the model is trained.

**Theorem 5.1.** For a particular model architecture, κ must be constant.

**Proof:** From Theorem 4.1 (Inverse Proportionality of Learning Rate), we established that for large datasets, the optimal learning rate follows:

$$\eta \propto \frac{1}{D}$$

meaning that as D increases, η scales inversely with D. Therefore,

$$\eta(x) = \frac{C}{D}$$

Substituting this into the integral:

$$\kappa = \int_0^D \frac{C}{D} dx$$

Which yields:

$$\kappa = \frac{C}{D} * D = C$$

For some constant C. Thus, C is exactly equal to κ, the total learning effort required for convergence.

The cumulative learning constant, κ, encapsulates the total step distance traveled by the model in parameter space to converge to the local minimum of the loss function. For a specific model architecture, κ depends on the properties of the loss landscape, including the distance to the minimum and the curvature of the loss surface, both of which are determined by the model's architecture and hyperparameters. While different datasets may influence the trajectory of the optimization process, the total step distance required for convergence remains constant for sufficiently large datasets because the overall shape of the loss landscape (as governed by the architecture) remains unchanged.

The theoretical analysis in these proofs is based on the assumption that the model is tested under a fixed seed and converges to the same local minimum across different runs. However, in the experimental setup, the model was initialized with a random seed, introducing stochastic variations in the optimization trajectory and convergence to different local minima. The assumption of a fixed seed is primarily used to ensure that the initial loss across different runs remains approximately identical, which holds in expectation under the assumption that the dataset is drawn from the same underlying distribution.

Furthermore, in a sufficiently complex loss landscape, it is reasonable to assume that the optimizer encounters a statistically equivalent structure of peaks, troughs, and curvature variations across different runs, regardless of the specific trajectory taken in weight space. This follows from the assumption that the loss function exhibits a degree of ergodicity, meaning that for sufficiently large datasets, the optimizer traverses a representative subset of the loss landscape. Under this assumption, the expected total optimization distance required for convergence remains approximately constant, as formalized by Theorem 5.1.

Despite this theoretical invariance in total optimization distance, experimental results indicate that increasing the dataset size while decreasing the learning rate (as dictated by the inverse proportionality relationship) consistently resulted in a lower final evaluation loss. The experimental results not only confirm the validity of these theorems but also demonstrate that the model does not converge to the same minimum across different runs. This property can be leveraged to efficiently determine an optimal learning rate using a significantly smaller dataset. By applying the established proportionality relationships, the optimal learning rate can then be scaled appropriately for training on a much larger dataset, allowing the model to still benefit from the increased data while maintaining computational efficiency.

We now consider a scenario with a scheduled decaying learning rate. Regardless of the specific schedule, the total step distance required to minimize the loss remains κ. This invariance arises because κ represents the total learning effort required to traverse the loss landscape to the minimum, which is determined solely by the architecture and not by the specific learning rate schedule. However, for this invariance to hold, the learning rate schedule must be chosen such that the optimization process converges to the minimum without overshooting or stagnation. Poorly designed schedules may lead to suboptimal convergence, but when properly optimized, they all yield the same cumulative step distance κ.

Theorem 5.1 has several implications. First, a more sophisticated model can have its datasets scaled down and the optimal learning rate can be found using much less computational power. Second, the optimal learning rate can be discovered using a constant learning rate. The constant κ can then be calculated and used to solve for the optimal learning rate on more sophisticated learning rate schedulers. For example, the data shown in figure (5) demonstrates that under this specific model architecture, at 13 epochs an optimal learning rate is around 0.000423. The learning constant for this is:

$$\ell * D = 0.000423 * (13 * 30000) \approx 165$$

The next experiment will involve a decaying learning rate in which the learning rate halves after each epoch. To solve for κ in this specific scenario, we can replace the integral with the geometric series

$$\eta D \sum_{n=1}^{13} \frac{1}{2^{n-1}} = 165$$

Which yields $\eta \approx 0.00275$. This represents the initial value the decaying learning rate should be set to in order to minimize loss. Figure (8) demonstrates different losses when varying the initial value of the decaying learning rate.

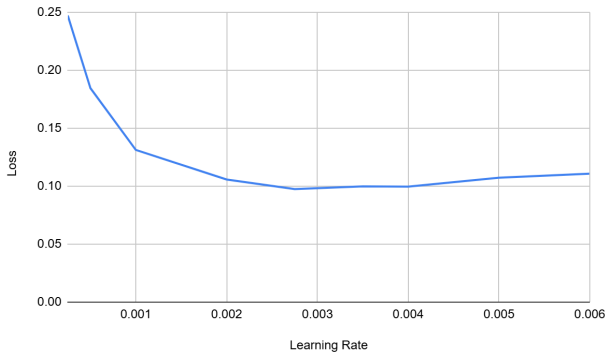

Figure 8. Loss vs. Decaying Learning Rate on 10 epochs using an Adam optimizer.

9 points were taken varying the learning rate from 0.000225 to 0.006. The optimal learning rate from these was in fact 0.00275. The same experiment was repeated for 5 epochs. The calculation for the optimal learning rate using the cumulative learning method was 0.00284. The results are shown in figure (9) below.

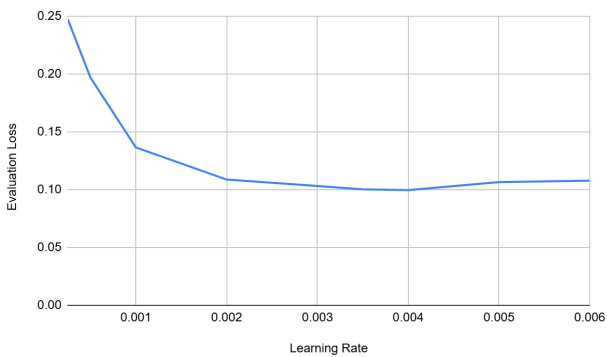

Figure 9. Loss vs. Decaying Learning Rate on 5 epochs using an Adam optimizer.

Here, the loss minimized at a learning rate of 0.004. However, this is likely an issue with large variance at lower amounts of data. It can be seen from figure (5) that point 13 fits nearly perfectly on the line of best fit, while point 5 varies. This experiment was done on 22 epochs. The graph visualizing the expected optimal learning rate vs. the optimal learning rate can be seen in Figure (10) below.

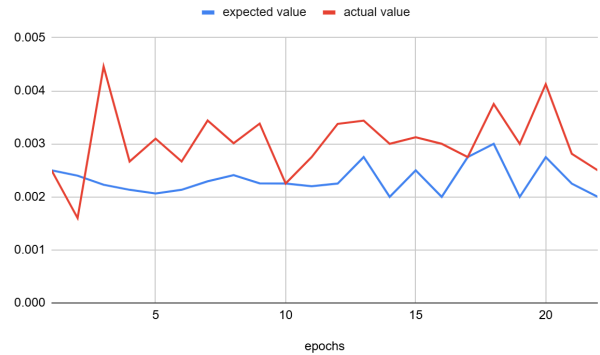

Figure 10. The expected optimal learning rate and the experimentally obtained optimal learning rate are both plotted vs. epochs using the Adam optimizer and a decaying learning rate schedule.

At epochs < 5, there appears to be a large variance in data. This is likely due to the lack of substantial data leading to testing inconsistency. Nonetheless, while the actual values seem far from the expected values, the coefficient of variation was calculated to be only 22.4%. This experiment was repeated with a decaying learning rate under a SGD optimizer. The results can be seen in figure (11).

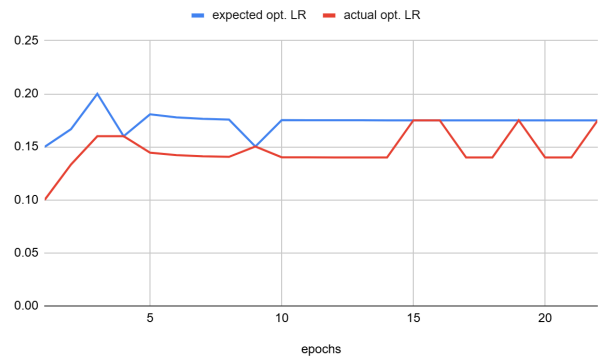

Figure 11. The expected optimal learning rate and the experimentally obtained optimal learning rate are both plotted vs. epochs using an SGD optimizer and a decaying learning rate schedule.

Figure (11) demonstrates that even when using SGD instead of Adam, the same correlation appears.

This test was also performed with a cyclical learning rate schedule instead of a decaying learning rate schedule. The cyclical learning rate tripled at every even epoch and returned to its original value at every odd epoch. The results from this experiment may be seen in figure (12).

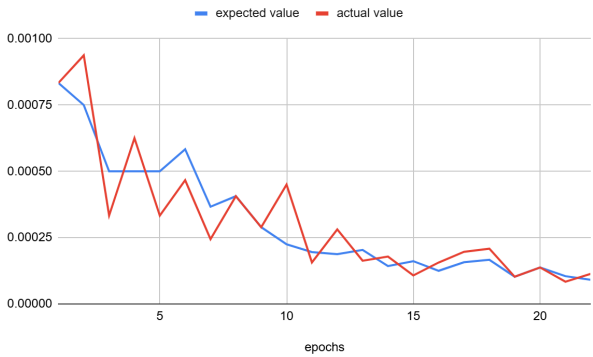

Figure 12. Expected learning rate and actual learning rate for an Adam optimizer under a cyclical learning rate schedule.

At epochs < 10, the results are very choppy. However, it can be seen that the optimal learning rate still converges to the expected value calculated using the learning rate constant under large datasets.

This same experiment was also done with an SGD optimizer. This can be seen in figure (13) below.

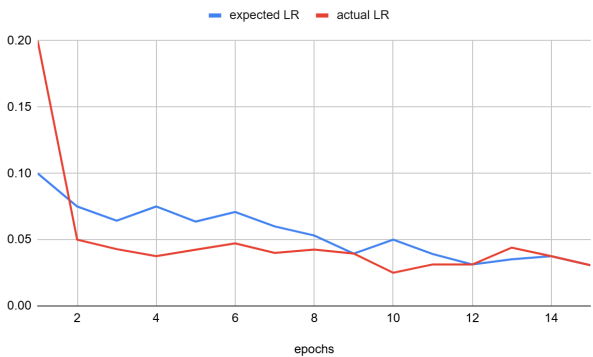

Figure 13. Expected learning rate and actual learning rate for an SGD optimizer under a cyclical learning rate schedule.

Once again, figure (13) demonstrates that for higher epochs there is a very close correlation between the expected optimal learning rate using the cumulative learning constant and the experimentally obtained optimal learning rate.

## 6. Conclusion

This work uncovers a fundamental inverse proportionality between the optimal learning rate and dataset size, a relationship that holds especially strong for both SGD and Adam optimizers and becomes even more pronounced as training data increases. This discovery provides a critical insight into scaling laws for deep learning, revealing a pattern that could reshape how learning rates are optimized in large-scale models.

Furthermore, we introduce the concept of a cumulative learning constant, a quantity that can be easily computed under a constant learning rate and leveraged to design more efficient and adaptive learning rate schedules. This breakthrough has the potential to streamline hyperparameter tuning, reducing the computational cost and inefficiencies associated with current learning rate selection methods.

The implications of this research extend far beyond classification tasks. Future work should explore even larger datasets and alternative architectures, such as reinforcement learning agents, generative models, and large language models, to determine the full extent of these findings. If validated across broader applications, this framework could lay the groundwork for a new era of automated, parameter-free learning rate optimization, ultimately pushing the boundaries of deep learning efficiency and scalability.

The principles established in this study have the potential to accelerate the adoption of automated learning rate selection in deep learning. By reducing reliance on manual tuning, this approach could be incorporated into future AutoML frameworks, enabling faster, more efficient training of large-scale models without exhaustive hyperparameter sweeps. Further research could validate its application to architectures such as transformers, reinforcement learning agents, and generative models, potentially setting the stage for fully automated, self-optimizing neural networks.